\documentclass[10pt,twocolumn,letterpaper]{article}

\usepackage{cvpr}              
\usepackage{graphicx}
\usepackage{booktabs}
\usepackage{cite}
\usepackage[colorlinks,linkcolor=red]{hyperref}
\usepackage{colortbl, color, xcolor}
\usepackage{amsmath}
\usepackage{mathtools}
\usepackage{amssymb}
\usepackage{multirow}

\definecolor{cvprblue}{rgb}{0.21,0.49,0.74}

\newcommand{\subtitle}[1]{{\noindent}{\textbf{#1}}}

\def\paperID{63} 
\def\confName{CVPR}
\def\confYear{2024}

\title{Improving Bracket Image Restoration and Enhancement with Flow-guided Alignment and Enhanced Feature Aggregation}

\author{
    Wenjie Lin\textsuperscript{\rm1} \qquad 
    Zhen Liu\textsuperscript{\rm1} \qquad 
    Chengzhi Jiang\textsuperscript{\rm1} \qquad 
    Mingyan Han\textsuperscript{\rm1} \qquad \\
    Ting Jiang\textsuperscript{\rm1} \qquad  
    Shuaicheng Liu\textsuperscript{\rm2,\rm1 $\dagger$}\\
    \textsuperscript{\rm1}Megvii Technology \ \textsuperscript{\rm2}University of Electronic Science and Technology of China \\
    \url{https://github.com/NygmaLin/IREANet}
}

\begin{document}
\maketitle
\begin{abstract}
In this paper, we address the Bracket Image Restoration and Enhancement (BracketIRE) task using a novel framework, which requires restoring a high-quality high dynamic range (HDR) image from a sequence of noisy, blurred, and low dynamic range (LDR) multi-exposure RAW inputs. To overcome this challenge, we present the IREANet, which improves the multiple exposure alignment and aggregation with a Flow-guide Feature Alignment Module (FFAM) and an Enhanced Feature Aggregation Module (EFAM). Specifically, the proposed FFAM incorporates the inter-frame optical flow as guidance to facilitate the deformable alignment and spatial attention modules for better feature alignment. The EFAM further employs the proposed Enhanced Residual Block (ERB) as a foundational component, wherein a unidirectional recurrent network aggregates the aligned temporal features to better reconstruct the results. To improve model generalization and performance, we additionally employ the Bayer preserving augmentation (BayerAug) strategy to augment the multi-exposure RAW inputs. Our experimental evaluations demonstrate that the proposed IREANet shows state-of-the-art performance compared with previous methods. 
\end{abstract}    
\section{Introduction}
\label{sec:intro}

Image restoration and enhancement (IRE), which aims to obtain high-quality image results from single or multiple degraded images, is a fundamental yet still challenging low-level vision problem. Compared to single-image restoration and enhancement~\cite{zamir2020cycleisp,zhang2017beyond,nah2017deep,zamir2021multi,eilertsen2017hdr,liu2020single,zou2023rawhdr,jiang2023low}, utilizing multi-frame information to solve these ill-posed problems is more practical and has shown greater potential~\cite{wang2021deep,Dudhane_2022_CVPR,perez2021ntire,bhat2022ntire}.

Existing multi-frame IRE methods can be broadly divided into two categories: BurstIRE and Multi-exposure IRE. BurstIRE methods exploit multiple consecutive frames of images with identical exposure to tackle tasks like denoising, deblurring, and high dynamic range (HDR) imaging. For instance, burst denoising~\cite{godard2018deep,mildenhall2018burst,rong2020burst,xia2020basis} leverages a burst capture of short-exposure to predict a single noise-free and unblurred image estimate by considering the fact that noise is independent across different frames. Burst deblurring~\cite{aittala2018burst,delbracio2015burst,wieschollek2017end} aggregates a burst of inputs into a single image that is both sharper and less noisy than
all the images in the burst. Techniques like HDR+~\cite{hasinoff2016burst} utilize raw bursts for more practical HDR reconstruction on commercial smartphones.

\begin{figure}[t]
  \centering
  \includegraphics[width=1.0\linewidth]{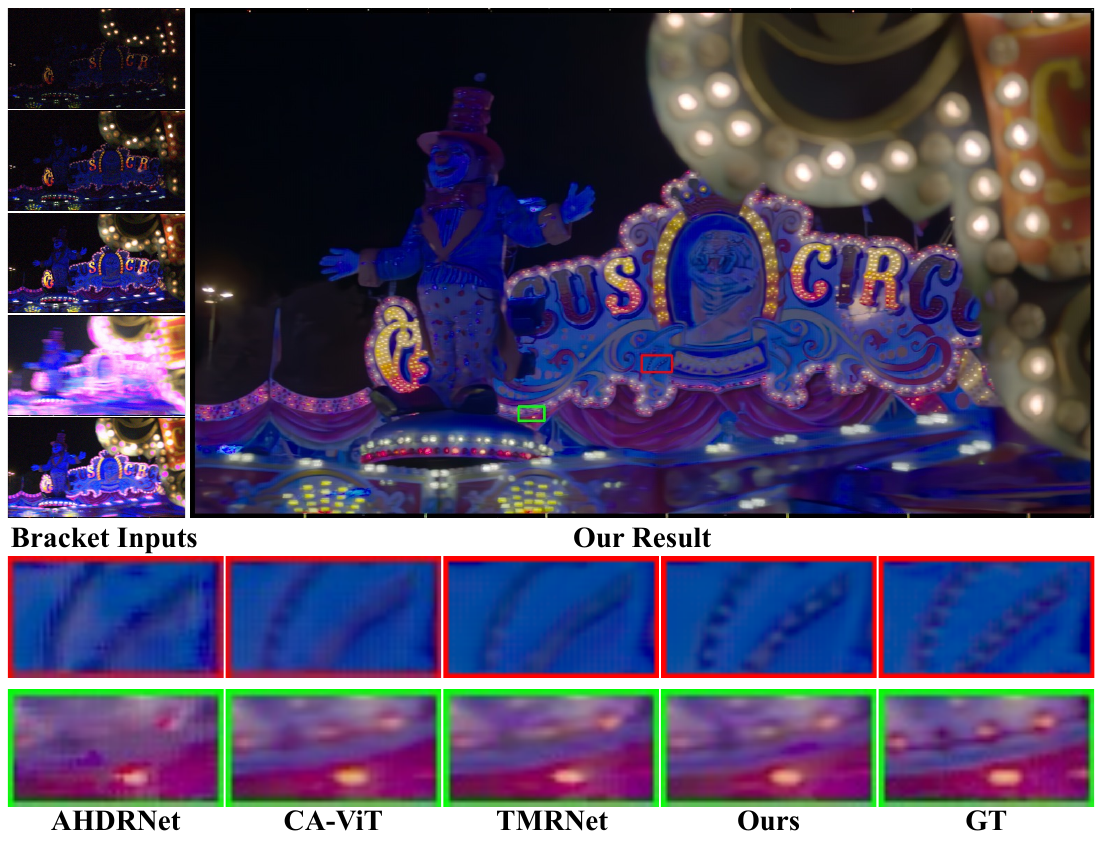}   
  \caption{Comparisons between the propsoed IREANet and other representative methods~\cite{yan2019attention,liu2022ghost,zhang2024bracketing}.  Our result is free of noise and blur while producing more details.}\label{fig:teaser}
\end{figure}

However, the limited scene information of burst inputs, e.g., dynamic range and noise level, hinders the further development of BurstIRE. Conversely, by benefiting from a wider dynamic range and richer levels of noise and motion blur across different exposures, Multi-exposure IRE has recently received more attention for being more efficient and practical in tasks such as HDR reconstruction~\cite{kalantari2017deep,wu2018deep,yan2019attention,niu2021hdr,liu2021adnet,liu2022ghost} and denoising~\cite{chi2023hdr,liu2023joint}. Motivated by the complementary nature of multi-exposure images in denoising, deblurring, and HDR reconstruction, the NTIRE2024 (New Trends in Image Restoration and Enhancement) introduced the BracketIRE challenge~\cite{ntire2024bracketing}, which aims to jointly perform denoising, deblurring, and HDR reconstruction tasks by utilizing multi-exposure images.

In this paper, we propose the IREANet, which improves the feature alignment and feature aggregation in the BracketIRE task. Specifically, the proposed IREANet mainly comprises a flow-guided feature alignment module (FFAM) and an enhanced feature aggregation module (EFAM). The proposed FFEM employs the optical flow between bracket exposures to guide the optimization of the deformable alignment and the spatial attention, enabling more effective alignment of the bracket inputs. Moreover, to enhance the stability and efficiency of the feature aggregation, we propose an Enhanced Residual Block (ERB), which enhances the nonlinearity of the vallina residual block by adding a $1 \times 1$ convolutional layer and a nonlinear activation layer. Through the incorporation of the proposed ERB as the foundational module, the EFAM achieves better convergence, thereby allowing for efficient aggregation of temporal features. Additionally, to preserve the correct Bayer pattern of the RAW images, we adopt a Bayer preserve augmentation (BayerAug) strategy to eliminate the issues of Bayer misalignment in the training data during augmentation. Based on the aforementioned components, the proposed IREANet is capable of producing visually pleasing, high-quality HDR results with clear content and enhanced details (as illustrated in Fig.~\ref{fig:teaser}).

Our main contributions can be summarized as follows: 
\begin{itemize}
  \item We propose a flow-guided feature alignment module (FFAM), which utilizes the optical flow between bracket images to guide the deformable alignment module and the spatial attention module, achieving better alignment performance.
  \item We propose an enhanced feature aggregation module (EFAM),  which uses the proposed enhanced residual block (ERB) as its foundational component, enabling more efficient aggregation of inter-frame features.
  \item We adapt the Bayer preserving augmentation (BayerAug) to raw bracket images for more effective augmentation, allowing the reconstruction of finer details. 
  \item Experimental results demonstrate that the proposed approach achieves better performance than the state-of-the-art methods, both quantitively and qualitatively.
\end{itemize}

\section{Related Work}
\label{sec:related_work}

\subtitle{Burst Image Restortation and Enhancement} Image enhancement and restoration based on burst inputs primarily focus on tasks such as image denoising~\cite{godard2018deep,mildenhall2018burst,rong2020burst,xia2020basis}, deblurring~\cite{aittala2018burst,delbracio2015burst,wieschollek2017end}, and super-resolution~\cite{bhat2021burst,Bhat_2023_ICCV,Luo_2021_CVPR,Bhat_2021_CVPR,luo2022bsrt}, which utilize the information from multiple consecutive frames with the same exposure to restore the target image. For instance, Godard \emph{et al.} demonstrated that burst inputs provide large improvements over deep learning single-frame denoising techniques~\cite{godard2018deep}. Delbracio \emph{et al.} introduced a fast method for merging a burst of images into a single image, which is both sharper and less noisy than all the images in the burst. Goutam \emph{et al.} recently introduced a method that leverages deep learning networks for burst super-resolution, which takes multiple noisy raw images as input and generates a single RGB image that is both denoised and super-resolved. Although BurstIRE methods have a clear advantage over single-image IRE, they still lack abundant scene information and priors, such as a broader dynamic range and richer levels of noise and motion blur.

\subtitle{Multi-exposure Image Restoration and Enhancement} Multi-exposure-based image enhancement and restoration primarily focus on HDR reconstruction, since images with different exposures can provide distinct details and dynamic range information. Kalantari \emph{et al.} first proposed a CNN-based approach for multi-exposure HDR imaging of dynamic scenes. Subsequently, a lot of CNN-based~\cite{wu2018deep,yan2019attention,niu2021hdr,prabhakar2020towards,prabhakar2021labeled,liu2021adnet} and Transformer-based~\cite{liu2022ghost,tel2023alignment,chen2023improving,song2022selective} methods have been proposed to enhance alignment and fusion performance in multi-exposure HDR reconstruction. Recently, several methods have been proposed that take noise and resolution into account, offering joint HDR reconstruction along with denoising or super-resolution~\cite{chi2023hdr,liu2023joint,tan2021deep}. In this paper, we consider more practical scenarios with severe noise and motion blur and propose a method capable of effectively performing joint HDR reconstruction, denoising, and deblurring.

\section{Method}
\label{sec:method}

\subsection{Overview}

\begin{figure*}[t]
  \centering
  \includegraphics[width=0.9\linewidth]{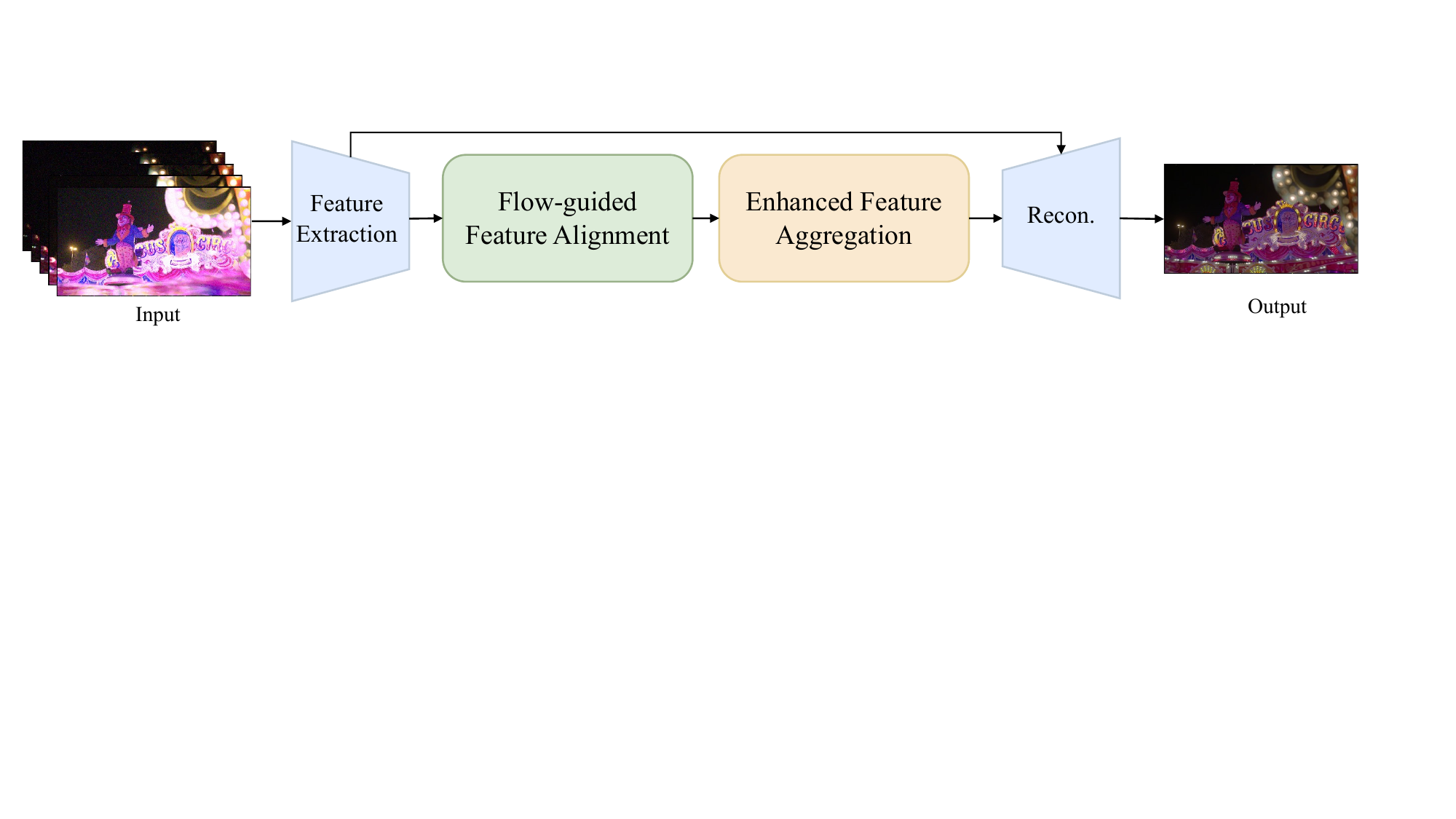}   
  \caption{Illustration of the overall framework of the proposed IREANet. As shown in Fig.~\ref{fig:pipeline}(a), the pipeline mainly consists of four components: feature extraction, feature alignment, feature aggregation, and reconstruction. Fig.~\ref{fig:pipeline}(b) illustrates the proposed flow-guided feature alignment module (FFAM). Fig.~\ref{fig:pipeline}(c) depicts the differences between the vallina residual block (RB) and the proposed enhanced residual block (ERB).}\label{fig:pipeline}
\end{figure*}

As illustrated in Fig.~\ref{fig:pipeline}(a), the overall pipeline of the proposed IREANet mainly consists of four components, i.e., feature extraction, flow-guided feature alignment, enhanced feature aggregation, and reconstruction.

Given the input bracket multi-exposure images $\{x_{i}\}^{N}_{i=1}$ ($x_{i}\in \mathbb{R}^{4\times H \times W}$, N is the number of input exposures, H, W is the height and width of the raw image), which are all 4-channels `RGGB' RAW images with noise, blur, and low dynamic range, the task of BracketIRE aims to reconstruction a clear and high dynamic range RAW image $I_{H}\in \mathbb{R}^{4\times H \times W}$. In our experiments, we take the shortest-exposure image $x_{1}$ as the reference frame $x_{ref}$ and the remaining images as non-reference frames.

\subsection{Network Structure}

\subtitle{Feature Extraction} Given the input images $\{x_{i}\}^{N}_{i=1}$, We first extract informative features $\{F_{i}\}^{N}_{i=1}$ through a feature extraction module, i.e., 
\begin{align}\label{eq:feature_extract}
    \{F_{i}\}^{N}_{i=1} = \mathcal{M}_{FE}(\{x_{i}\}^{N}_{i=1}).
\end{align}
The feature extraction module consists of several residual blocks. In the meanwhile, we send the input images to the SpyNet to obtain the optical flows which are calculated between each no-reference frame $\{x_{i}\}^{N}_{i=2}$ and the reference frame $x_{ref}$:
\begin{align}\label{eq:flow_extract}
    \{f_{i}\}^{N}_{i=2} = \mathcal{M}_{SpyNet}(x_{i}, x_{ref}), i\in 2, 3, ..., N
\end{align}
The non-reference features $\{F_{i}\}^{N}_{i=2}$ are then warped by the estimated optical flows, generating the warped features:
\begin{align}\label{eq:warp_feature}
    \hat{F}_{i} = \mathcal{W}(F_{i}, f_{i}),
\end{align}
where $\mathcal{W}$ denotes the wrapping operator. More precisely, the $\hat{F}_{1}$ is the same as the reference feature.

\subtitle{Flow-guided Feature Alignment} Inspired by BasicVSR++~\cite{chan2022basicvsr++} and BSRT~\cite{liu2022ghost}, we employ the optical flows to guide the feature alignment. Specifically, we propose a flow-guided feature alignment module (FFAM), which is designed as a dual-branch architecture to perform feature alignment. The first branch takes the extracted features and optical flows as input, with the optical flows being considered a rough alignment prior, to guide the DCNs in learning more accurate offsets for flow-guided deformable alignment (FDA). 
\begin{align}\label{eq:deformable_align}
    \{FF_{i}\}^{N}_{i=1} = \mathcal{M}_{FDA}(\{F_{i}\}^{N}_{i=1}, \{\hat{F}_{i}\}^{N}_{i=1}, \{f_{i}\}^{N}_{i=2}).
\end{align}

The second branch performs spatial feature attention (SFA) on the warped features, obtaining spatial attention features:
\begin{align}\label{eq:spatial_attention}
    \{SF_{i}\}^{N}_{i=1} = \mathcal{M}_{SFA}(\{\hat{F}_{i}\}^{N}_{i=1}).
\end{align}
The spatial attention mechanism has been proven to effectively reduce noise and undesired contents caused by foreground object movement~\cite{yan2019attention,liu2022ghost}. The flow-guided aligned features $\{FF_{i}\}^{N}_{i=1}$ and the spatial attention features $\{SF_{i}\}^{N}_{i=1}$, serving as complementary features, are then combined through element-wise addition to acquire the final aligned features $\{\check{F}_{i}\}^{N}_{i=1}$.

\subtitle{Enhanced Feature Aggregation} Similar to TMRNet~\cite{zhang2024bracketing}, a unidirectional recurrent network is utilized to aggregate temporal features. However, through our experiments, we discovered that the aggregation module using original residual blocks tends to induce training instabilities and fails to converge effectively. Consequently, we introduce an enhanced feature aggregation module (EFAM), which incorporates the proposed enhanced residual block (ERB) as basic components. The proposed ERB increases the network's nonlinearity and enables better convergence, thus allowing for more effective aggregation of informative temporal features. This process can be formulated as:
\begin{align}\label{eq:feature_aggregation}
    AF = \mathcal{M}_{EFAM}(\{\check{F}_{i}\}^{N}_{i=1}).
\end{align}

\subtitle{Reconstruction} Finally, the aggregated features $AF$ are fed into a reconstruction network composed of several residual blocks, generating the final high-quality HDR result. We also adopt the skip connection to stabilize the training process, i.e., 
\begin{align}\label{eq:recon}
    I_{H} = \mathcal{M}_{Recon}(AF) + F_{1}.
\end{align}

\begin{figure}[t]
   \centering
   \includegraphics[width=1.0\linewidth]{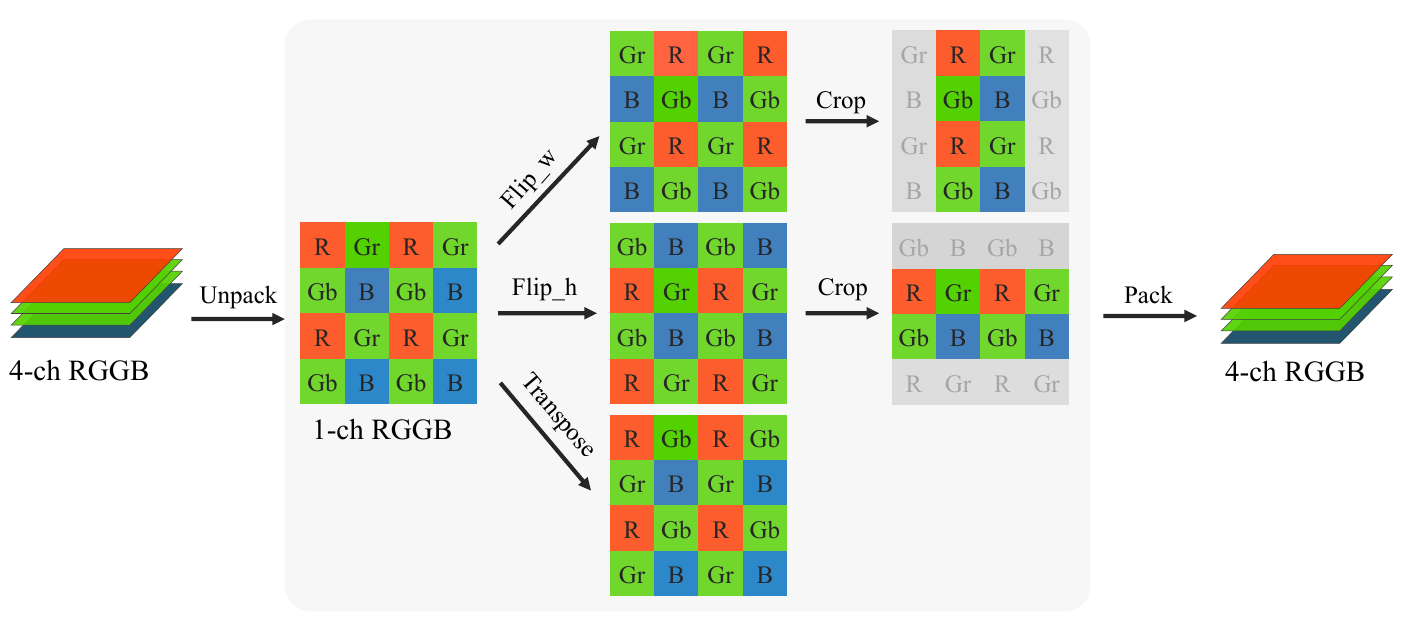}   
   \caption{Illustration of the adopted Bayer preserving augmentation (BayerAug).}\label{fig:bayer_aug}
 \end{figure}

\begin{table*}[t]
    \centering
    \caption{Quantitative comparisons of our method with several representative multi-frame IRE methods on the BracketIRE dataset. We report the widely used metrics PSNR, SSIM, and LPIPS.}
    \label{tab:quantative_results}
    \resizebox{1.0\linewidth}{!}{
      \begin{tabular}{
      c
      >{\centering\arraybackslash}p{1.2cm}
      >{\centering\arraybackslash}p{1.4cm}
      >{\centering\arraybackslash}p{1.4cm}
      >{\centering\arraybackslash}p{1.4cm}
      >{\centering\arraybackslash}p{1.4cm}
      >{\centering\arraybackslash}p{1.6cm}
      >{\centering\arraybackslash}p{1.4cm}
      >{\centering\arraybackslash}p{1.4cm}
      >{\centering\arraybackslash}p{1.6cm}}
    \toprule
    \multirow{2}{*}{Method} &MFIR & Burstormer & AHDRNet & HDRGAN & CA-ViT & SCTNet & Kim et al. & TMRNet & IREANet \\ 
                          &  \cite{bhat2021deep} & \cite{dudhane2023burstormer}   & \cite{yan2019attention}   & \cite{niu2021hdr}&\cite{liu2022ghost}& \cite{tel2023alignment}& \cite{kim2023joint}& \cite{zhang2024bracketing}& Ours \\ \midrule
    PSNR $\uparrow$   & 33.92    & 37.01   &  36.32   &  35.07  &  36.54  & 36.90 & 37.93  & 38.19 &  \textbf{39.78} \\
    SSIM $\uparrow$   & 0.9026   & 0.9454  &  0.9273  &  0.9157 &  0.9341 & 0.9437& 0.9452 & 0.9488&  \textbf{0.9556}\\ 
    LPIPS $\downarrow$ & 0.196    & 0.127   &  0.154   &  0.177  &  0.127  & 0.120 & 0.115  & 0.112 &  \textbf{0.102} \\
    \bottomrule
    \end{tabular}
}
\end{table*}

\begin{figure*}[h]
   \centering
   \includegraphics[width=1.0\linewidth]{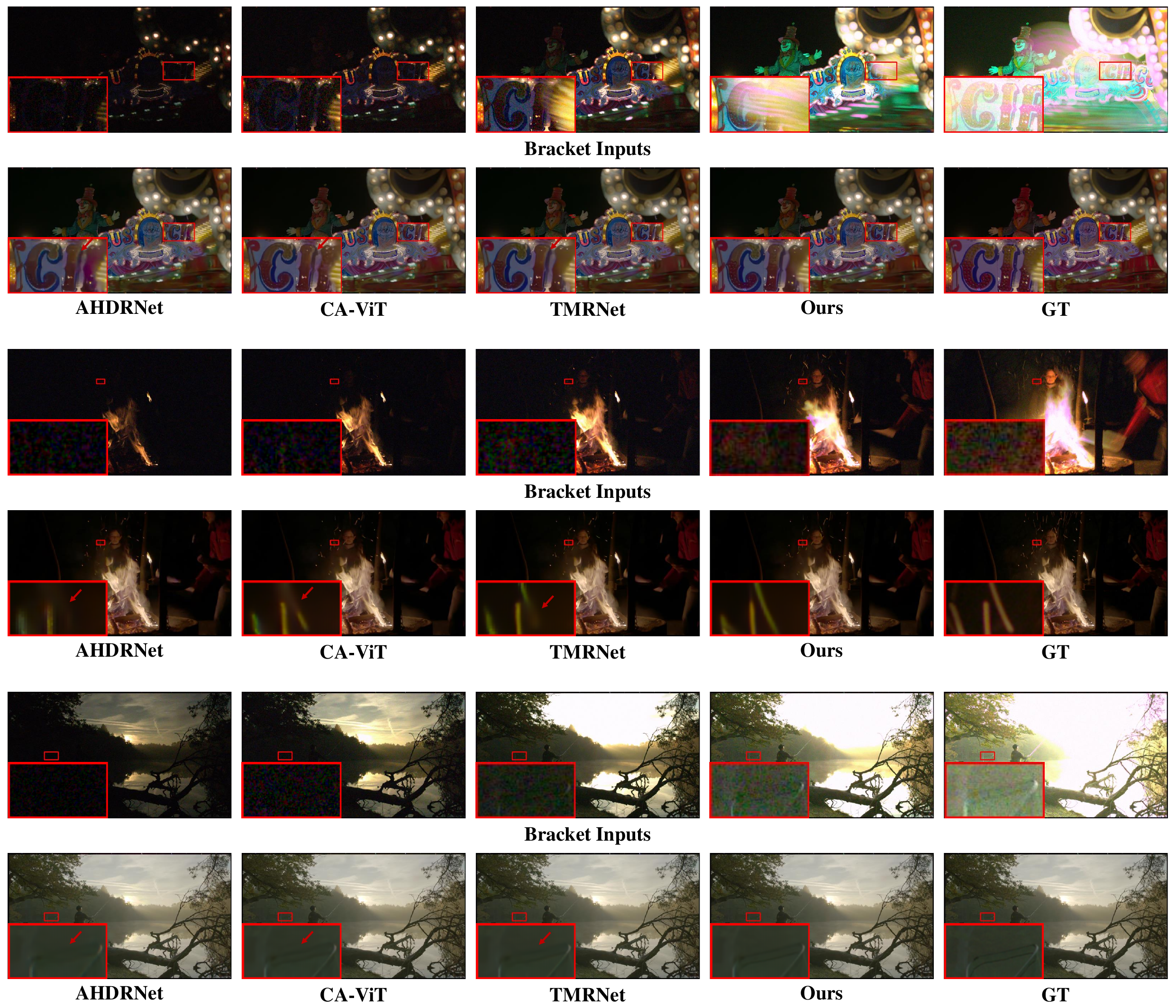}   
   \caption{Qualitative Comparison with existing state-of-the-art methods AHDRNet~\cite{yan2019attention}, CA-ViT~\cite{liu2022ghost}, and TMRNet~\cite{zhang2024bracketing}. Our approach can effectively restore high-quality image details from multiple exposure inputs. }\label{fig:method_cmp}
 \end{figure*}

\subsection{Training Strategy}

\subtitle{Bayer Preserving Augmentation} As demonstrated in ~\cite{liu2019learning}, employing augmentation methods designed for RGB images on raw images is error-prone. Therefore, we utilize the Bayer preserving augmentation (BayerAug) to augment the training data. The primary process of the BayerAug is shown in Figure ~\ref{fig:bayer_aug}. As can be seen, a vertical or horizontal flip switches the Bayer pattern from RGGB to GRBG or GBRG, which may lead to errors during the training process, thereby affecting the details of the reconstruction results. To preserve the correct Bayer pattern during augmentation, we combine flipping and cropping together. Taking vertical flipping as an example, after flipping an image, we apply horizontal cropping to ensure that the Bayer pattern remains correct after augmentation.

\subtitle{Loss Function}
Inspired by existing methods~\cite{kalantari2017deep,yan2019attention,liu2022ghost}, we compute the loss in the tonemapped domain as HDR images are commonly viewed after tonemapping. We use the $\mu$-law function as the tonemapping operator:
\begin{align}
\mathcal{T}(x)=\frac{\log (1+\mu x)}{1+\mu},
\end{align}
where $\mathcal{T}(x)$ is the tonemapped HDR image. In this paper, we set $\mu$ to 5000. In our experiments, we utilize the widely adopted $l_{1}$ loss as the loss function, i.e.,
\begin{align}
\mathcal{L}=\|\mathcal{T}(I_{\mathrm{GT}})-\mathcal{T}(I_{\mathrm{H}})\|_{1},
\end{align}
where $I_{\mathrm{GT}}$ and  $I_{\mathrm{H}}$ denote the ground-truth HDR image and the reconstructed image, respectively.

\section{Experiment}
\label{sec:experiment}

\subsection{Dataset and Implementation Details}
We train and evaluate our method on the Bracekt dataset~\cite{ntire2024bracketing} provided by the NTIRE2024 Bracketing Image Restoration and Enhancement Challenge. It contains 1045 scenes in total. We select 100 scenes as the validation set and keep the remaining as the training set. Each scene consists of five LDR images with their corresponding HDR ground truth. Each LDR image contains degradations like ill-exposure, noise, and blur.
During the training stage, we first crop the input LDR images to $128\times128$ patches. We use Bayer pattern augmentation with a random combination of vertical flip,  horizontal flip, and rotation. The network is optimized by an AdamW optimizer with a cosine learning rate scheduler. The initial learning rate is 0.0001 and the decay rate is 0.01. 

Our experiments are implemented in PyTorch and trained on 4 NVIDIA 2080Ti GPUs with a batch size of 8. We train the model from scratch for 1000 epochs, and the whole training costs about 176 hours. We select the best model using the PSNR score calculated on our validation set when the training reaches plateaus. The entire testing process is conducted on a single Tesla P40 GPU. We test our model with the full-size ($5 \times 4 \times 960 \times 540$) of input, and it takes about 6.1s and 15G memory for each sample. We compute the PSNR, SSIM, and LPIPS scores as the testing metrics with the images on tonemapped domain.

\subsection{Results and Analysis}
To demonstrate the superiority of our proposed IREANet, we compare it with existing state-of-the-art methods, which include two burst-based methods (i.e., MFIR~\cite{bhat2021deep} and Burstomer~\cite{dudhane2023burstormer}) and six bracket-based methods (i.e., AHDRNet~\cite{yan2019attention}, HDR-GAN~\cite{niu2021hdr}, CA-ViT~\cite{liu2022ghost}, SCTNet~\cite{tel2023alignment}, Kim \emph{et al.}~\cite{kim2023joint}, and TMRNet~\cite{zhang2024bracketing}). For fair comparisons, the quantitative results of previous work are borrowed from TMRNet~\cite{zhang2024bracketing} except AHDRNet~\cite{yan2019attention} and CA-ViT~\cite{liu2022ghost}, which we retrained in the BracketIRE~\cite{ntire2024bracketing} dataset following the same settings as our method to obtain both quantitative and qualitative results. 

\subtitle{Quantitative results} The quantitative results are listed in Table \ref{tab:quantative_results}. As can be seen, the proposed IREANet outperforms both the burst-based and bracket-based approaches by a large margin, achieving a 1.59dB gain than the second-best method TMRNet~\cite{zhang2024bracketing}.

\subtitle{Qualitative results} Fig. \ref{fig:method_cmp} illustrates the qualitative results compared with the existing state-of-the-art methods. For each scene, the first row shows the input bracket images, and the second row presents the compared tonemapped results. As shown, in the first scene, the input brackets exhibit ill exposure along with lens flare, making it difficult for recent SOTA methods to recover the texture of the image. In the second and third scenes, the input multi-exposure images suffer from heavy noise and motion blur. These factors cause other methods to encounter issues such as loss of image details and lack of sharpness in these scenes. On the contrary, our proposed IREANet reconstructs high-quality HDR results with realistic image details, which further demonstrates the effectiveness of our method.

\begin{table}[t]
    \centering
    \caption{Quantitative results of the ablation studies.}
    \label{tab:ablation}
    \resizebox{1.0\linewidth}{!}{
      \begin{tabular}{
      c
      >{\centering\arraybackslash}p{1.2cm}
      >{\centering\arraybackslash}p{1.6cm}
      >{\centering\arraybackslash}p{1.2cm}
      >{\centering\arraybackslash}p{1.2cm}
      >{\centering\arraybackslash}p{1.2cm}
      >{\centering\arraybackslash}p{1.2cm}}
    \toprule
    ---      & Bayer~Aug.  &  EFAM       &  FFAM       & PSNR     & SSIM  \\ \midrule
    Baseline &             &             &             & 38.19    & 0.9488\\
    Variant1 & $\checkmark$&             &             & 39.02    & 0.9515\\ 
    Variant2 & $\checkmark$& $\checkmark$&             & 39.54    & 0.9541\\ \midrule
    IREANet  & $\checkmark$& $\checkmark$& $\checkmark$& 39.78    & 0.9556\\
    \bottomrule
    \end{tabular}
    }
\end{table}


\begin{figure}[t]
  \centering
  \includegraphics[width=1.0\linewidth]{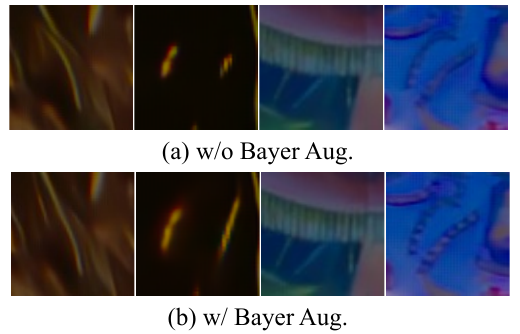}   
  \caption{Qualitative results of our ablation study on the Bayer preserve augmentation.}\label{fig:result_bayer_aug}
\end{figure}

\subsection{Ablation Study}
To verify the effectiveness of the proposed IREANet, we conduct ablation studies of our contributions and analyze the results on the test set provided by NTIRE2024 Bracketing Image Restoration and Enhancement Challenge. We compare our method with three variants as follows:
\begin{itemize}
    \item \textbf{Baseline}. Our method takes the TMRNet~\cite{zhang2024bracketing} as our baseline, which is one of the SOTA methods in recent BracketIRE works.
    \item \textbf{Variant1}. This variant applies augmentations with random flip and rotation in a Bayer augmentation manner. 
    \item \textbf{Variant2}. This variant replaces the basic enhanced residual convolution blocks (ERB) in the backbone of the baseline with our proposed enhanced feature aggregation module (EFAM).
    \item \textbf{IREANet}. The entire network architecture of the proposed IREANet, which contains an enhanced feature aggregation module (EFAM) and flow-guided feature alignment module (FFAM).
\end{itemize}

As shown in Table \ref{tab:ablation}, variant 1 only applies the Bayer augmentation on TMRNet in the training process. 
Compared with the baseline model and vanilla augmentation, training with the augmentations in a Bayer augmentation manner shows better performance. The variant1 has a 0.83dB gain of PSNR. The main reason can be concluded that there are differences between data augmentation for Bayer raw images and RGB images. Employing an appropriate image augmentation on bayer raw images can make the augmented data closer to the real data distribution, leading to an improvement in network training effectiveness. The qualitative comparison results between bayer augmentation and the baseline can be seen in Fig. \ref{fig:result_bayer_aug}. When we apply the bayer augmentation method on bayer raw images to train the network, the generated images gain richer textures and clearer content.

\begin{figure}[t]
  \centering
  \includegraphics[width=0.85\linewidth]{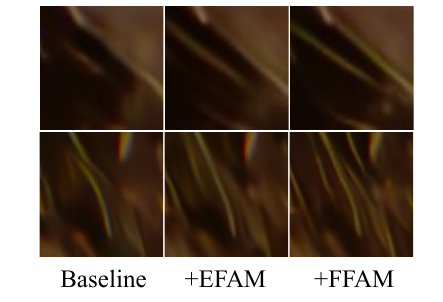}   
  \caption{Qualitative comparison of our ablation study on the proposed EFAM and FFAM.}\label{fig:result_ablation_net}
\end{figure}

We also conduct experiments to explore the effectiveness of our proposed module. As shown in Table \ref{tab:ablation}, the variant2 where the ERBs in the network backbone of baseline are replaced with EFAM, showing 0.52dB improvement in PSNR compared to variant1. Additionally, incorporating FFAM on variant 2 gains another 0.24dB improvement, demonstrating the effectiveness of our proposed modules. The qualitative comparison of the ablation experiments on the modules we proposed can be seen in Fig. \ref{fig:result_ablation_net}. Employing EFAM on baseline structure obtains richer textures, and further adding FFAM improves the sharpness of image textures, thus validates the conclusion mentioned earlier.
\section{Conclusion}
\label{sec:conclusion}
In this paper, we have introduced a novel framework (namely IREANet) for the task of Bracket Image Restoration and Enhancement (BracketIRE), which requires utilizing a sequence of noisy, blurred, and low dynamic range multi-exposure RAW inputs to generate high-quality high dynamic range (HDR) images. The model incorporates a Flow-guide Feature Alignment Module (FFAM) and an Enhanced Feature Aggregation Module (EFAM) to improve feature alignment and aggregation. Additionally, we have employed a Bayer-preserving augmentation strategy to improve the generalization capabilities while maintaining the correct Bayer pattern. Experimental results have shown that our IREANet achieves state-of-the-art performance, producing high-quality HDR images with clear content and enhanced details. The proposed method has also achieved second place in the NTIRE 2024 BracketIRE Challenge.

{
    \small
    \bibliographystyle{ieeenat_fullname}
    \bibliography{main}
}


\end{document}